\documentclass[runningheads]{llncs}
\usepackage{graphicx}
\usepackage{amsfonts}
\usepackage{amsmath}
\usepackage{hyperref}
\hypersetup{colorlinks=true}
\usepackage{cite}
\usepackage{multirow}
\usepackage{booktabs}
\usepackage{pifont}
\usepackage{makecell}
\usepackage{xcolor}
\usepackage[super]{nth}
\usepackage[misc,geometry]{ifsym}
\newcommand{\cmark}{\text{\ding{51}}}

\begin{document}
\title{ICAL: Implicit Character-Aided Learning for Enhanced Handwritten Mathematical Expression Recognition}
\titlerunning{HMER with Implicit Character-Aided Learning}
%
\author{Jianhua Zhu\inst{1}\orcidID{0009-0000-3982-2739} \and
Liangcai Gao\inst{1}(\Letter) \and
Wenqi Zhao\inst{1}}
\authorrunning{J. Zhu et al.}
%

\institute{Wangxuan Institute of Computer Technology, Peking University, Beijing, China \\
\email{zhujianhuapku@pku.edu.cn}\\
\email{gaoliangcai@pku.edu.cn}\\
\email{wenqizhao@stu.pku.edu.cn}
}

\maketitle              
\begin{abstract}
Significant progress has been made in the field of handwritten mathematical expression recognition, while existing encoder-decoder methods are usually difficult to model global information in \LaTeX.  Therefore, this paper introduces a novel approach, Implicit Character-Aided Learning (ICAL), to mine the global expression information and enhance handwritten mathematical expression recognition. Specifically, we propose the Implicit Character Construction Module (ICCM) to predict implicit character sequences and use a Fusion Module to merge the outputs of the ICCM and the decoder, thereby producing corrected predictions. By modeling and utilizing implicit character information, ICAL achieves a more accurate and context-aware interpretation of handwritten mathematical expressions. Experimental results demonstrate that ICAL notably surpasses the state-of-the-art(SOTA) models, improving the expression recognition rate (ExpRate) by 2.25\%/1.81\%/1.39\% on the CROHME 2014/2016/2019 datasets respectively, and achieves a remarkable 69.06\% on the challenging HME100k test set. We make our code available on the GitHub.\footnote{\url{https://github.com/qingzhenduyu/ICAL}}
\end{abstract}
\section{Introduction}
\begin{figure}[htbp]
	\centering
	\includegraphics[scale=0.5]{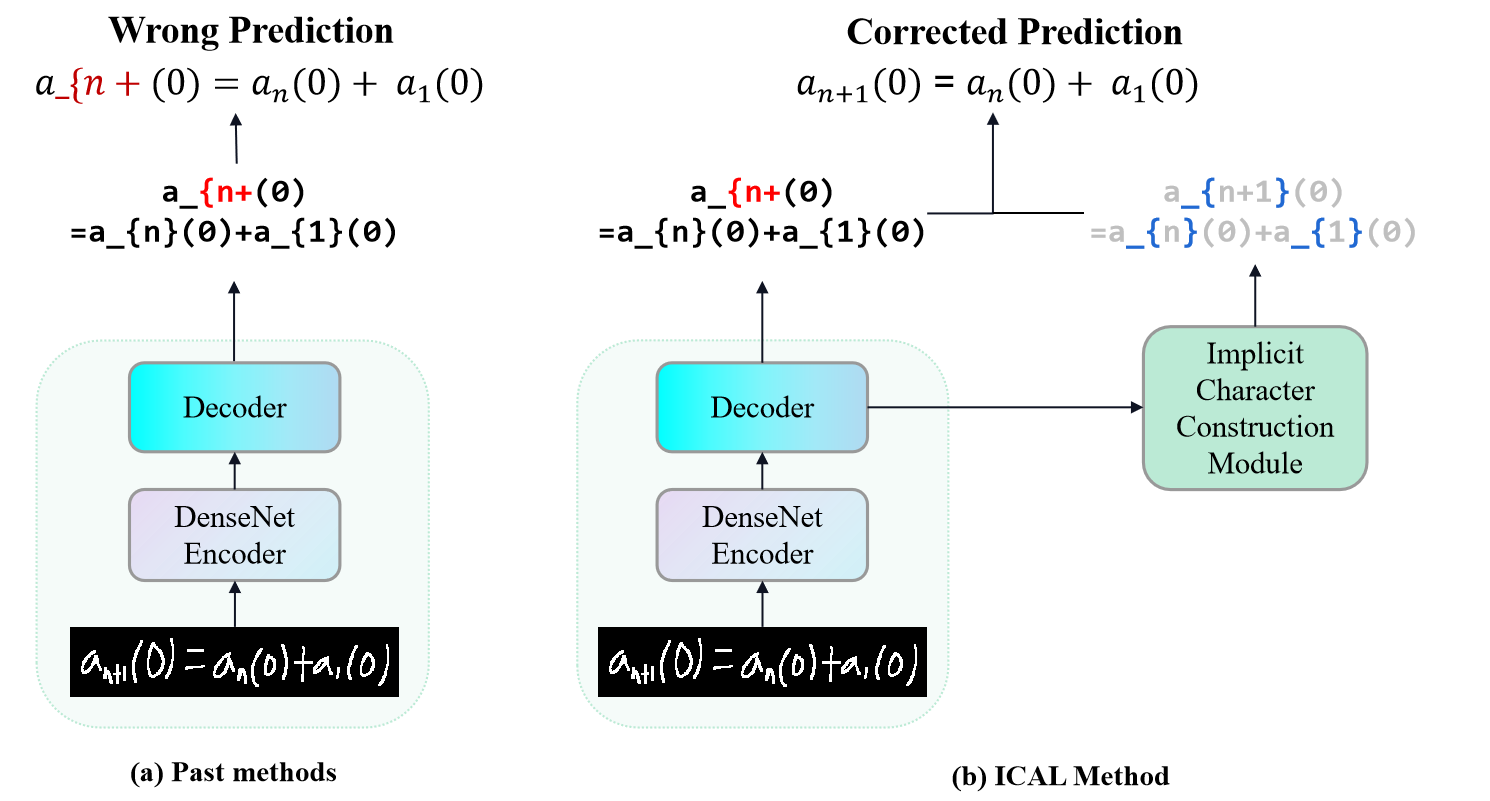}
	\caption{(a)Illustration of past method which uses DenseNet Encoder and RNN/Transformer Decoder. (b) Our ICAL method aided by implicit character learning. The characters highlighted in red signify inaccuracies in the prediction, whereas the blue highlights denote implicit characters.}
	\label{fig:overview}
\end{figure}

The Handwritten Mathematical Expression Recognition (HMER) task involves taking an image of a handwritten mathematical expression as input and having the model predict the corresponding \LaTeX. The HMER task has a wide range of applications, such as being used for intelligent grading of mathematical assignments, building online grading systems, and improving the efficiency of online education. Therefore, how to improve the recognition accuracy of handwritten mathematical expressions has become a hot topic in previous works.

Due to the diversity of handwriting and the two-dimensional structure of mathematical expressions, HMER is highly challenging. Compared to the recognition of handwritten text in natural language, the HMER task requires not only the prediction of explicit characters (i.e., characters that are directly represented when written by hand) but also implicit characters, such as ``\verb|^|", ``\verb|_|", ``\verb|{|", and ``\verb|}|", which is necessary to achieve a complete description of a two-dimensional mathematical expression. Past methods~\cite{zhang2018multi, zhao2021handwritten, zhao2022comer} based on encoder-decoder models often extract image features through the encoder, while the decoder aligns visual and textual features and predicts the \LaTeX. However, they often lack modeling of the global information of the expression, which in turn fails to correct prediction errors made by the decoder, as shown in the figure~\ref{fig:overview} (a):

In this paper, we utilize the task of predicting implicit characters (i.e., ``\verb|^|", ``\verb|_|", ``\verb|{|", and ``\verb|}|") to assist the decoder in modeling the global information of \LaTeX, which can further correct the output of the decoder and improve recognition performance. To this end, we propose an Implicit Character Construction Module, capable of modeling the sequence of implicit characters from the output of the Transformer Decoder~\cite{vaswani2017attention}. This global information is then passed to a subsequent Fusion Module which integrates the information with the output of the Transformer decoder to achieve a more accurate prediction of the \LaTeX\ sequence.

In this work, the main contributions of our work are summarized as follows:
\let\labelitemi\labelitemii
\begin{itemize}
  \item We introduced the Implicit Character Construction Module(ICCM) to model implicit character information, which can effectively utilize the global information in \LaTeX.
  \item We proposed the Fusion Module to aggregate the output of the ICCM, thereby correcting the prediction of the Transformer Decoder.
  \item Experimental results indicate that the ICAL method surpasses previous state-of-the-art methods and achieves expression recognition rate (ExpRate) of 60.63\%, 58.79\%, and 60.51\% on the CROHME 2014~\cite{mouchere2014icfhr}/2016~\cite{mouchere2016icfhr2016}/2019~\cite{mahdavi2019icdar} test sets, respectively, and an ExpRate of 69.06\% on the HME100K test set~\cite{yuan2022syntax}.
\end{itemize}

\section{Related Work}

\subsection{Traditional Methods}
In traditional handwritten mathematical expression recognition, the process mainly involves symbol recognition and structural analysis. Symbol recognition requires segmenting and identifying individual symbols within expressions, utilizing techniques like pixel-based segmentation proposed by OKAMOTO et al~\cite{599029}. and further refined by HA et al~\cite{ha1995understanding}. into recursive cropping methods. These approaches often depend on predefined thresholds. For symbol classification, methods like Hidden Markov Models (HMM)\cite{winkler1996hmm,kosmala1999line,alvaro2014recognition}, Elastic Matching\cite{chan1998elastic, vuong2010towards}, and Support Vector Machines (SVM)\cite{keshari2007hybrid} have been used, where HMMs allow for joint optimization without explicit symbol segmentation, though at a high computational cost.

Structural analysis employs strategies like the two-dimensional Stochastic Context-free Grammar (SCFG)\cite{chou1989recognition} and algorithms such as Cocke-Younger-Kasami (CKY)\cite{sakai1961syntax} for parsing, albeit slowly due to the complexity of two-dimensional grammar. Faster parsing has been achieved with Left-to-right Recursive Descent and Tree Transformation methods\cite{zanibbi2002recognizing}, the latter describing the arrangement and grouping of symbols into a structured tree for parsing. Some approaches bypass two-dimensional grammar altogether, using Define Clause Grammar (DCG) \cite{chan2000efficient}and Formula Description Grammar\cite{toyota2006structural} for one-dimensional parsing, highlighting the challenges in designing comprehensive grammars for the diverse structures of mathematical expressions.

\subsection{Deep Learning Methods}
In recent years, encoder-decoder-based deep learning models have become the mainstream framework in the field of HMER. Depending on the architecture of the decoder, past deep learning approaches can be categorized into methods based on Recurrent Neural Networks (RNNs)~\cite{zhang2017watch, zhang2018multi, ding2021encoder, zhang2018track, wu2019image,wu2020handwritten, truong2020improvement, bian2022handwritten, li2022counting, zhang2020treedecoder, wu2022tdv2, yuan2022syntax} and those based on the Transformer model~\cite{zhao2021handwritten, zhao2022comer,zhang2023general}. Furthermore, based on the decoding strategy, these methods can be divided into those based on sequence decoding~\cite{zhang2017watch, zhang2018multi, ding2021encoder, zhang2018track, wu2019image,wu2020handwritten, truong2020improvement, bian2022handwritten, li2022counting,zhao2021handwritten, zhao2022comer,zhang2023general} and those based on tree-structured decoding
~\cite{zhang2020treedecoder, wu2022tdv2, yuan2022syntax}.

\noindent \textbf{RNN-based methods} In 2017, Zhang et al. proposed an end-to-end deep learning model, WAP~\cite{zhang2017watch}, to address the problem of HMER. The encoder part of the model is a fully convolutional neural network similar to VGGnet~\cite{simonyan2014very}. The decoder part uses a GRU~\cite{cho2014learning} model to generate the predicted \LaTeX\ sequence from the extracted visual features. The WAP not only avoids issues caused by inaccurate symbol segmentation but also eliminates the need for manually predefined \LaTeX\ syntax, thereby becoming a benchmark model for subsequent deep learning methods. Following WAP, Zhang et al. further proposed the DenseWAP~\cite{zhang2018multi}, which replaces the VGGnet with the DenseNet~\cite{huang2017densely}. Subsequent work has commonly adopted the DenseNet as the backbone network for the encoder. The CAN model~\cite{li2022counting} introduces a Multi-Scale Counting Module, utilizing a symbol counting task as an auxiliary task to be jointly optimized with the expression recognition task.

\noindent \textbf{Transformer-based methods}
To alleviate the issue of unbalanced output and fully utilize bidirectional language information, BTTR~\cite{zhao2021handwritten} adopts a bidirectional training strategy on top of the Transformer-based decoder. Following this, CoMER ~\cite{zhao2022comer} incorporates coverage information~\cite{tu2016modeling, zhang2017watch} into Transformer Decoder, introducing an Attention Refinement Module. This module utilizes the attention weights from the Multi-head Attention mechanism within the Transformer decoder to compute the coverage vector, all while maintaining the characteristic of parallel decoding. Based on CoMER, the GCN~\cite{zhang2023general} incorporates extra symbol categorization information , utilizing the General Category Recognition Task as a supplementary task for joint optimization with the HMER task, resulting in a notable performance. However, the category recognition task introduced by GCN requires manual construction of symbol categories and is limited to specific datasets.

\noindent \textbf{Tree-based decoding methods} \LaTeX, as a markup language, can be easily parsed into a tree-like expression due to the influence of delimiters such as brackets. Therefore, by leveraging the inherent two-dimensional structure of mathematical expressions , models can provide certain interpretability for the prediction process. Zhang et al. proposed the DenseWAP-TD ~\cite{zhang2020treedecoder}, which replaces the GRU decoder that directly regresses the \LaTeX\ sequence with a decoder based on a two-dimensional tree structure. The TDv2 model~\cite{wu2022tdv2}, during training, uses different transformation methods for the same \LaTeX\ string, weakening the context dependency and endowing the decoder with stronger generalization capabilities. The SAN model~\cite{yuan2022syntax} converts the \LaTeX\ sequence into a parsing tree and designs a series of syntactic rules to transform the problem of predicting \LaTeX\ sequences into a tree traversal process. Additionally, SAN introduces a new Syntax-Aware Attention Module to better utilize the syntactic information in \LaTeX.

\section{Methodology}
\begin{figure}[htbp]
	\centering
	\includegraphics[scale=0.4]{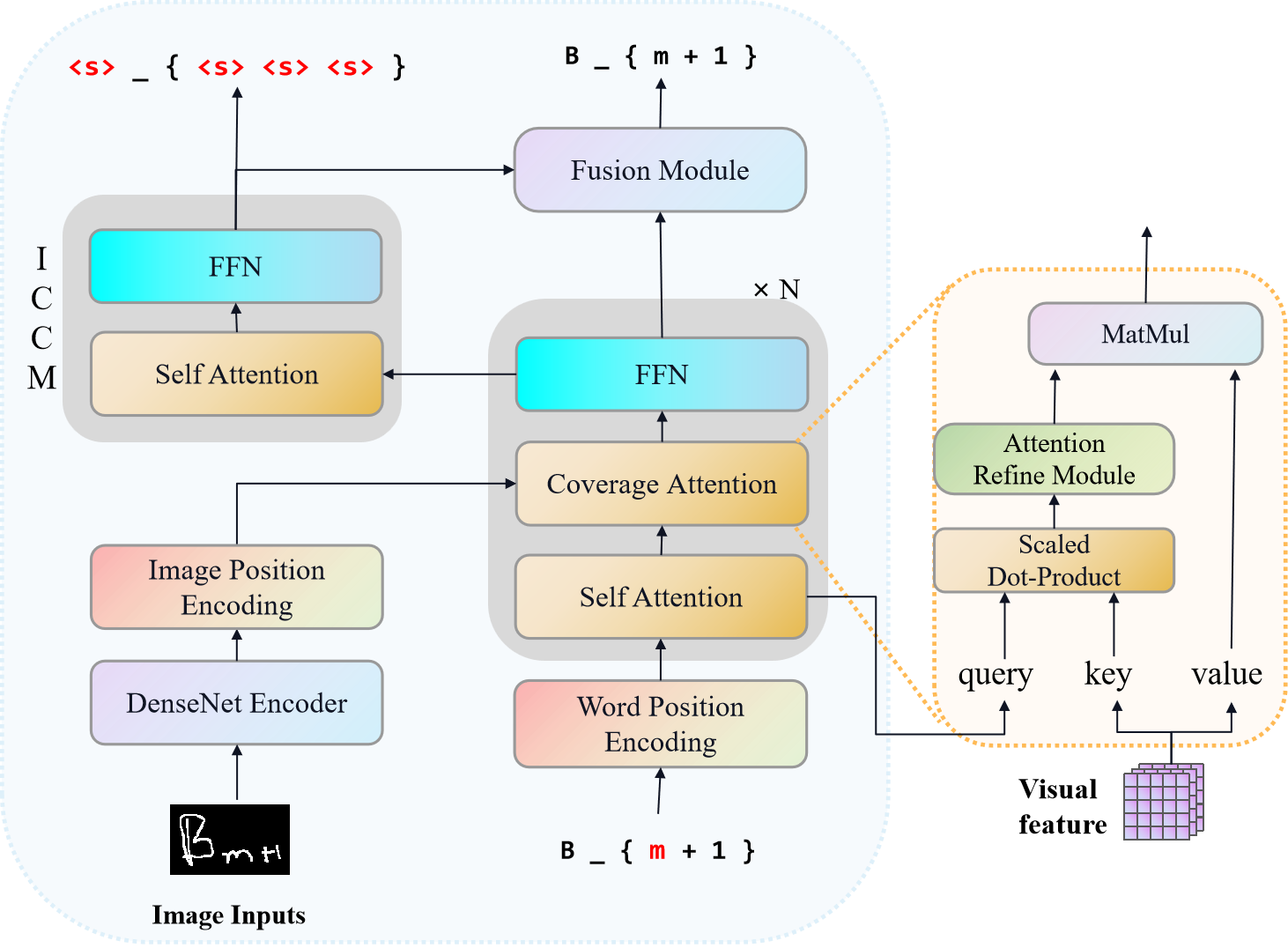}
	\caption{The architecture of ICAL model (left) and Coverage Attention (right). To simplify the illustration, we have condensed the depiction of bidirectional training in the figure.}
	\label{fig:model}
\end{figure}

In this section, we will elaborate in detail on the model structure of the ICAL we proposed, as shown in the figure~\ref{fig:model}.
In Section~\ref{sec:cnn}, we will briefly introduce the DenseNet~\cite{huang2017densely} used in the encoder part. In Section~\ref{sec:decoder}, we will introduce the Transformer decoder that adopts coverage attention mechanism~\cite{vaswani2017attention,zhao2022comer}, which can alleviate the lack of coverage problem. In Sections~\ref{sec:implicit} and~\ref{sec:fusion}, we will introduce the Implicit Character Construction Module(ICCM) proposed in this paper and the Fusion Module that integrates implicit character information. Finally, in Section~\ref{sec:loss}, we will discuss how the implicit character loss and fusion loss are introduced while employing a bidirectional training strategy~\cite{zhao2021handwritten}.

\subsection{Visual Encoder} \label{sec:cnn}
Similar to most of the previous work~\cite{zhang2018multi, ding2021encoder, zhang2018track, wu2019image,wu2020handwritten, truong2020improvement, bian2022handwritten, li2022counting,zhao2021handwritten, zhao2022comer,zhang2023general, zhang2020treedecoder, wu2022tdv2, yuan2022syntax}, we continue to use DenseNet~\cite{huang2017densely} as the visual encoder to extract features from the input images. 

DenseNet consists of multiple dense blocks and transition layers. Within each dense block, the input for each layer is the concatenated outputs from all preceding layers, enhancing the flow of information between layers in this manner. The transition layers reduce the dimensions of the feature maps using $1 \times 1$ convolutional kernels, decreasing the number of parameters and controlling the complexity of the model. 

For an input grayscale image of size $1 \times H_0 \times W_0$, the output visual feature $\mathbf V_{feature} \in \mathbb{R}^{D \times H \times W}$. The ratios of $H$ to $H_0$ and $W$ to $W_0$ are both 1/16. In this work, a $1 \times 1$ convolutional layer is used to adjust the number of channels $D$ to $d_{\text{model}}$, aligning it with the dimension size of the Transformer decoder.

\subsection{Transformer Decoder with ARM} \label{sec:decoder}
In the decoder part, we employ a Transformer Decoder with Attention Refine Module(ARM) as the decoder~\cite{vaswani2017attention,zhao2022comer}, which mainly consists of three modules: Self-Attention Module, Coverage Attention Module, and Feed-Forward Network.

\noindent \textbf{Self-Attention Module} The self-attention module functions as a core component of the standard Transformer decoder, utilizing a masked multi-head attention mechanism. This module processes a set of queries ($\mathbf Q$), keys ($\mathbf K$), and values ($\mathbf V$) to produce an output that leverages both dot-product attention and a multi-head strategy for processing information. For each head within the multi-head attention framework, linear transformations are applied to $\mathbf Q$, $\mathbf K$, and $\mathbf V$ using transformation matrices $\mathbf W_i^{q}$, $\mathbf W_i^{k}$, and $\mathbf W_i^{v}$, respectively, where $i$ denotes the index of the head.

The mechanism first calculates a scaled dot-product attention score $\mathbf E_i$ for the $i$-th head by multiplying the transformed queries and keys and then scaling the result by the square root of the dimension of the key vectors $\mathbf K$ to avoid large values that could hinder softmax computation:

\begin{equation}
\mathbf E_{i} = \frac{(\mathbf{QW}_{i}^{q})(\mathbf{KW}_i^{k})^T}{\sqrt{d_{k}}},
\end{equation}

A softmax function is then applied to these scores to obtain the attention weights $A_i$:

\begin{equation}
\mathbf A_{i} = \operatorname{softmax}(\mathbf E_{i}),
\end{equation}

These weights are used to compute a weighted sum of the values, producing the output $H_i$ for each head:

\begin{equation}
\mathbf H_{i} = \mathbf A_{i}(\mathbf{VW}_i^{v}),
\end{equation}

Finally, the outputs of all heads are concatenated and linearly transformed to produce the final output of the multi-head attention module:

\begin{equation}
\operatorname{MultiHeadAttention}(\mathbf{Q}, \mathbf K, \mathbf V) = \operatorname{Concat}(\mathbf H_{0}, \mathbf H_{1}, \ldots, \mathbf H_{\text{head}-1})\mathbf W^{o}.
\end{equation}

Since the decoder performs decoding in an autoregressive manner, the prediction of the current symbol depends on past predicted symbols and input visual information. To avoid the decoder obtaining information about future symbols when predicting the current symbol and to ensure parallelism, Self-Attention Module uses a masked lower-triangular matrix to constrain the information that the self-attention module can access at the current step.

\noindent \textbf{Coverage Attention Module}
The CoMER model, without affecting the parallelism of the Transformer, introduces coverage attention commonly used in RNNs ~\cite{zhang2017watch} into the Transformer decoder by improving the Cross Attention module. Within the Cross Attention module of the CoMER model, an Attention Refine Module (ARM) is incorporated. By utilizing alignment information from the previous layer and the current layer, it refines the current attention weights $A_{i}$, enabling the decoder to faithfully convert the text structure from visual features to corresponding \LaTeX\ text. The update formulas for the attention weights in the j-th layer of the decoder, denoted as $\hat{A}^{j}$, are as follows in Equations \ref{math:E} and \ref{math:A}. 

\begin{equation}
    \mathbf{\hat{E}}^{j} = \operatorname{ARM}(E^{j},\mathbf{\hat{A}}^{j-1}) \label{math:E},
\end{equation}

\begin{equation}
    \mathbf{\hat{A}}^{j} = \operatorname{softmax}(\mathbf{\hat{E}}^{j}) \label{math:A}.
\end{equation}

\noindent \textbf{Feed-Forward Network}
The feed-forward neural network consists of two linear layers and the ReLU non-linear activation function. For input $\mathbf X$ from Coverage Attention Module, it is as follows:
\begin{equation}
    \mathbf E_{feature} = \operatorname{FFN}(\mathbf{X}) = \operatorname{ReLU}(\mathbf{XW}_{0}+\mathbf b_{0})\mathbf{W}_{1} + \mathbf b_{1}.
\end{equation}

\subsection{Implicit Character Construction Module} \label{sec:implicit}
For the target \LaTeX\ sequence, we only retain its implicit characters, namely ``\verb|^|", ``\verb|_|", ``\verb|{|", and ``\verb|}|", replacing other characters with newly constructed ones to form a sequence corresponding to the implicit characters. For example, for the target \LaTeX\ sequence \verb| B _ { m + 1 } |, the corresponding implicit character sequence is \verb| <space> _ { <space> <space> <space> }|. 

The output of the Transformer Decoder with ARM serves as the input to our Implicit Character Construction Module (ICCM), which consists of a layer of masked self-attention and a Feed-Forward Network (FFN) layer. Consequently, the output of the ICCM, $\mathbf{I}_{\text{feature}}$, is calculated as follows:
\begin{equation}
 \mathbf I_{feature} = \operatorname{FFN}(\operatorname{Self Attention}(\mathbf E_{feature})).
\end{equation}

\subsection{Fusion Module}\label{sec:fusion}
To integrate the information learned by the ICCM, we introduce a weighted adjustment strategy based on the attention mechanism, capable of aggregating the output of the ICCM, thereby correcting the prediction results of the Transformer Decoder. 

The Fusion Module assimilates inputs from both the ICCM's output, denoted as $\mathbf{I}_{\text{feature}} \in \mathbb{R}^{B \times T \times d_{\text{model}}}$, and the Transformer Decoder's output, denoted as $\mathbf{E}_{\text{feature}} \in \mathbb{R}^{B \times T \times d_{\text{model}}}$. The attention weights $\mathbf f_{att}$ are calculated as follows:

\begin{equation}
 \mathbf f_{att} = \sigma(w_{att}(\text{Concat}(\mathbf E_{feature}, \mathbf I_{feature})),
\end{equation}
where a linear layer $w_{att}$ maps the concatenate feature matrix back to the original dimension $d_{\text{model}}$, and attention weights $f_{att}$ are computed through a sigmoid activation function $\sigma$.

Finally, $f_{att}$ are used to perform a weighted fusion of the two features, resulting in the final output feature for prediction.
\begin{equation}
 \mathbf F = \mathbf f_{att} \odot \mathbf E_{feature} + (1 - \mathbf f_{att}) \odot \mathbf I_{feature},
\end{equation}
where $\odot$ indicates element-wise multiplication.

\subsection{Loss Function}\label{sec:loss}
To alleviate the issue of unbalanced output, we adhere to the bidirectional training strategy used in BTTR and CoMER, necessitating the computation of losses in both directions within the same batch. 

The total loss function is divided into three parts:
\begin{equation}
\mathcal{L} = \mathcal{L}_{Initial} + \mathcal{L}_{Implicit} + \mathcal{L}_{Fusion},
\end{equation}
where both $\mathcal{L}_{\text{Initial}}$ and $\mathcal{L}_{\text{fusion}}$ are standard cross-entropy loss functions. 

$\mathcal{L}_{\text{Initial}}$ calculates the loss using the Transformer decoder's predicted probabilities and the ground truth, consistent with the approach used in CoMER. $\mathcal{L}_{\text{fusion}}$ uses the predicted probabilities outputted by the Fusion Module and the ground truth. 

Since we construct the implicit character sequence directly from the \LaTeX\ sequence, keeping its length consistent with the original \LaTeX\ sequence, the constructed target implicit character sequence exhibits an imbalance in the occurrence frequency between \verb|<space>| and implicit characters. To address this, we use a weighted cross-entropy loss function to calculate the loss for implicit characters, as follows:

\begin{equation}
\mathcal{L}_{\text{implicit}} = -\sum_{i=1}^{N} \sum_{t=1}^{T_i} w_{y_{i,t}} \cdot \log(\hat{y}_{i,t}),
\end{equation}
where $N$ represents the batch size, $T_i$ is the length of the $i$-th sequence, $y_{i,t}$ is the ground truth token at position $t$ in the $i$-th sequence, $\hat{y}_{i,t}$ is the predicted probability of the correct token at position $t$ in the $i$-th sequence, and $w_{y_{i,t}}$ is the weight associated with the ground truth token $y_{i,t}$.

The weight $w_{y_{i,t}}$ for each token is dynamically adjusted based on the occurrence frequency of the token within the entire batch of sequences. The adjustment is made using a logarithmic function to ensure a smooth transition of weights across different frequencies:
\begin{equation}
w_{y_{i,t}} = 1.0 + \log\left(1 + \frac{1}{f_{y_{i,t}} + \epsilon}\right),
\end{equation}
where $f_{y_{i,t}}$ is the frequency of the token in the target sequences and $\epsilon$ is set to $1e-6$.

\section{Experiments}
\subsection{Dataset}
The CROHME dataset includes data from the Online Handwritten Mathematical Expressions Recognition Competitions (CROHME)~\cite{mouchere2014icfhr, mouchere2016icfhr2016, mahdavi2019icdar} held over several years and is currently the most widely used dataset for handwritten mathematical expression recognition. The training set of the CROHME dataset consists of 8,836 samples, while the CROHME 2014~\cite{mouchere2014icfhr}/2016~\cite{mouchere2016icfhr2016}/2019~\cite{mahdavi2019icdar} test sets contain 986, 1147, and 1199 samples respectively. In the CROHME dataset, each handwritten mathematical expression is stored in InkML format, recording the trajectory coordinates of the handwritten strokes. Before training, we converts the handwritten stroke trajectory information in the InkML files into grayscale images, and then carries out model training and testing.

The HME100K dataset~\cite{yuan2022syntax} is a large-scale collection of real-scene handwritten mathematical expressions. It contains 74,502 training images and 24,607 testing images, making it significantly larger than similar datasets like CROHME. Notably, it features a wide range of real-world challenges such as variations in color, blur, and complex backgrounds, contributed by tens of thousands of writers. With 249 symbol classes, HME100K offers a diverse and realistic dataset for developing advanced handwritten mathematical expression recognition systems.
\subsection{Evaluation Metrics}
The Expression Recognition Rate (ExpRate) is the most commonly used evaluation metric for handwritten mathematical expression recognition. It is defined as the percentage of expressions that are correctly recognized out of the total number of expressions. Additionally, we use the metrics “$\leq 1$ error” and “$\leq 2$ error” to describe the performance of the model when we tolerate up to 1 or 2 token prediction errors, respectively, in the \LaTeX\ sequence.
\subsection{Implementation Details}
We employs DenseNet~\cite{huang2017densely} as the visual encoder to extract visual features from expression images. The visual encoder utilizes 3 layers of DenseNet blocks, with each block containing 16 bottleneck layers. Between every two DenseNet blocks, a transition layer is used to reduce the spatial dimensions and the channel count of the visual features to half of their original sizes. The dropout rate is set at 0.2, and the growth rate of the model is established at 24. 

We  employ a 3-layer Transformer Decoder with ARM~\cite{vaswani2017attention, zhao2022comer} as the backbone of the decoder, with a model dimension ($d_{\text{model}}$) of 256 and head number set to 8. The dimension of the feed-forward layer is set to 1024, and the dropout rate is established at 0.3. The parameter settings for the Attention Rectifying Module (ARM) are consistent with those of CoMER, with the convolutional kernel size set to 5. Additionally, the parameters of masked self-attention and FFN in Implicit Character Construction Module(ICCM) are consistent with the aforementioned Transformer Decoder.

During the training phase, we utilize Mini-batch Stochastic Gradient Descent(SGD) to learn the model parameters, with weight decay set to 1e-4 and momentum set at 0.9. The initial learning rate is established at 0.08. We also adopt ReduceOnPlateau as the learning rate scheduler, whereby the learning rate is reduced to 25\% of its original value when the ExpRate metric ceases to change. When trained on the CROHME dataset, the CROHME 2014 test set~\cite{mouchere2014icfhr} is used as the validation set to select the model with the best performance. During the inference phase, we employ the approximate joint search method previously used in BTTR~\cite{zhao2021handwritten} to predict the output. 
\subsection{Comparison with State-of-the-art Methods}
Table ~\ref{tb:crohme_sota} presents the results on the CROHME dataset. To ensure fairness in performance comparison and considering that different methods have used various data augmentation techniques, many of which have not been disclosed, we have limited our comparison to results without the application of data augmentation. Given the relative small size of the CROHME dataset, we conducted experiments with both the baseline CoMER and the proposed ICAL model using five different random seeds (7, 77, 777, 7777, 77777) under the same experimental conditions. The reported results are the averages and standard deviations of these five experiments.

It is noteworthy that while GCN ~\cite{zhang2023general} has attained impressive results on CROHME 2016~\cite{mouchere2016icfhr2016} and 2019~\cite{mahdavi2019icdar}, its performance benefits from the additional introduction of category information, while ICAL constructs the implicit character sequence directly from the \LaTeX\ sequence, eliminating the need for manually constructing additional category information. Consequently, we present the performance of the GCN solely for reference purposes and exclude it from direct comparisons. The CoMER model represents the current state-of-the-art(SOTA) method; however, CoMER\cite{zhao2022comer} did not disclose their results without data augmentation in the original paper. Therefore, we have reproduced the results of CoMER without data augmentation using their open-source code, denoted by $\dagger$ in the table~\ref{tb:crohme_sota}. 

As shown in table~\ref{tb:crohme_sota}, our method achieved the best performance across all metrics. Our method outperforms CoMER by 2.25\%/1.81\%/1.39\% on the CROHME 2014, 2016, and 2019 datasets, respectively. Across all metrics, the ICAL method achieves an average improvement of 1.6\% over the CoMER method. The experimental results on CROHME dataset prove the effectiveness of our method. 

We also conducted experiments on the challenging and real-world dataset HME100K, as shown in table~\ref{tb:hme100k_sota}. We have replicated the performance of CoMER on the HME100K dataset, and our method surpasses the state-of-the-art(SOTA) by 0.94\%, reaching an impressive 69.06\%. The outstanding experimental performance on the HME100K dataset, which is more complex, larger, and more realistic compared to CROHME, further proves the superior generalization ability and effectiveness of the ICAL method.

\begin{table*}[t]
    \centering
    \renewcommand\arraystretch{0.8}
    \caption{\textbf{Performance comparison on the CROHME dataset.} We compare expression recognition rate (ExpRate) between our model and previous state-of-the-art models on the CROHME 2014/2016/2019 test sets. \textbf{None of the methods used data augmentation to ensure a fair comparison.} We denote our reproduced results with $\dagger$. The symbol $*$ signifies the inclusion of supplementary information. All the performance results are reported in percentage (\%).}
    \label{tb:crohme_sota}
    \vspace{0.5em}
    \resizebox{\textwidth}{!}{
    \begin{tabular}{c|ccc|ccc|ccc}
    
    \toprule
    \multirow{2}{*}{\textbf{Method}} & \multicolumn{3}{c|}{\textbf{CROHME 2014}} & \multicolumn{3}{c|}{\textbf{CROHME 2016}} & \multicolumn{3}{c}{\textbf{CROHME 2019}} \\
    & ExpRate$\, \uparrow$ & $\leq 1\uparrow$ & $\leq 2\uparrow$ & ExpRate$\, \uparrow$ & $\leq 1\uparrow$ & $\leq 2\uparrow$ & ExpRate$\, \uparrow$ & $\leq 1\uparrow$ & $\leq 2\uparrow$ \\ 
    
    \midrule
    
    WAP & 46.55 & 61.16 & 65.21 & 44.55 & 57.10 & 61.55 & - & - & - \\
    DenseWAP & 50.1 & - & - & 47.5 & - & - & - & - & - \\
    DenseWAP-MSA &  52.8 &  68.1 & 72.0 & 50.1 & 63.8 & 67.4 & 47.7 & 59.5 & 63.3 \\
    TAP$^{*}$ &  48.47 &  63.28 & 67.34 & 44.81 & 59.72 & 62.77 & - & - & - \\
    PAL & 39.66 & 56.80 & 65.11 & - & - & - & - & - & - \\
    PAL-v2 & 48.88 & 64.50 & 69.78 & 49.61 & 64.08 & 70.27 & - & - & - \\
    WS-WAP & 53.65 & - & - & 51.96 & 64.34 & 70.10 & - & - & - \\
    ABM & 56.85 & 73.73 & 81.24 & 52.92 & 69.66 & 78.73 & 53.96 & 71.06 & 78.65 \\
    CAN-DWAP & 57.00 & 74.21 & 80.61 & 56.06 & 71.49 & 79.51 & 54.88 & 71.98 & 79.40 \\ 
    CAN-ABM & 57.26 & 74.52 & 82.03 & 56.15 & 72.71 & 80.30 & 55.96 & 72.73 & 80.57 \\

    \midrule
    
    DenseWAP-TD & 49.1 & 64.2 & 67.8 & 48.5 & 62.3 & 65.3 & 51.4 & 66.1 & 69.1 \\
    TDv2 & 53.62 & - & - & 55.18 & - & - & 58.72 & - & - \\
    SAN& 56.2 & 72.6 & 79.2 & 53.6 & 69.6 & 76.8 & 53.5 & 69.3 & 70.1 \\ 
    
    \midrule
    
    BTTR & 53.96 & 66.02 & 70.28 & 52.31 & 63.90 & 68.61 & 52.96 & 65.97 & 69.14 \\
    GCN$^{*}$ & 60.00 & - & - & 58.94 & - & - & 61.63 & - & - \\
    CoMER$^\dagger$ & 58.38$^{\pm 0.62}$ & 74.48$^{\pm 1.41}$  & 81.14$^{\pm 0.91}$ & 56.98$^{\pm 1.41}$ & 74.44$^{\pm 0.93}$  & 81.87$^{\pm 0.73}$ & 59.12$^{\pm 0.43}$ & 77.45$^{\pm 0.70}$ & 83.87$^{\pm 0.80}$ \\
    \textbf{ICAL} & \textbf{60.63}$^{\pm 0.61}$ & \textbf{75.99}$^{\pm 0.77}$ & \textbf{82.80}$^{\pm 0.40}$ & \textbf{58.79$^{\pm 0.73}$} & \textbf{76.06}$^{\pm 0.37}$ & \textbf{83.38}$^{\pm 0.16}$ & \textbf{60.51$^{\pm 0.71}$} & \textbf{78.00}$^{\pm 0.66}$ & \textbf{84.63}$^{\pm 0.45}$ \\
    \bottomrule
    
    \end{tabular}
    }
\end{table*}

\begin{table}[htbp]
	\renewcommand\arraystretch{0.8}
    \centering
    \caption{\textbf{Performance comparison on the HME100K dataset.} We compare our proposed ICAL with previous models on HME100K. We denote our reproduced results with $\dagger$. All the performance results are reported in percentage (\%).}
    \label{tb:hme100k_sota}
    \vspace{0.5em}
    
    \begin{tabular}{c|ccc}
    
    \toprule
    
    \multirow{2}{*}{\textbf{Method}} & \multicolumn{3}{c}{\textbf{HME100K}} \\
    & ExpRate$\, \uparrow$ & $\leq 1\uparrow$ & $\leq 2\uparrow$ \\
    
    \midrule
    
    DenseWAP & 61.85 & 70.63 & 77.14 \\
    DenseWAP-TD & 62.60 & 79.05 & 85.67 \\
    ABM & 65.93 & 81.16 & 87.86 \\
    SAN & 67.1 & - & - \\
    CAN-DWAP & 67.31 & 82.93 & 89.17 \\
    CAN-ABM& 68.09 & 83.22 & 89.91 \\
    \midrule
    BTTR & 64.1 & - & - \\
    CoMER$^\dagger$ & 68.12 & 84.20 & 89.71 \\

    \textbf{ICAL} & \textbf{69.06$^{\pm 0.16}$} & \textbf{85.16$^{\pm 0.13}$} & \textbf{90.61}$^{\pm 0.09}$ \\
    
    \bottomrule
    	
    \end{tabular}
    
\end{table}
\begin{table}[htbp]
	\renewcommand\arraystretch{0.8}
    \centering
    \caption{\textbf{Performance comparison on the HME100K dataset.} We compare our proposed ICAL with previous models on HME100K. We denote our reproduced results with $\dagger$. All the performance results are reported in percentage (\%).}
    \label{tb:hme100k_sota}
    \vspace{0.5em}
    
    \begin{tabular}{c|ccc}
    
    \toprule
    
    \multirow{2}{*}{\textbf{Method}} & \multicolumn{3}{c}{\textbf{HME100K}} \\
    & ExpRate$\, \uparrow$ & $\leq 1\uparrow$ & $\leq 2\uparrow$ \\
    
    \midrule
    
    DenseWAP & 61.85 & 70.63 & 77.14 \\
    DenseWAP-TD & 62.60 & 79.05 & 85.67 \\
    ABM & 65.93 & 81.16 & 87.86 \\
    SAN & 67.1 & - & - \\
    CAN-DWAP & 67.31 & 82.93 & 89.17 \\
    CAN-ABM& 68.09 & 83.22 & 89.91 \\
    \midrule
    BTTR & 64.1 & - & - \\
    CoMER$^\dagger$ & 68.12 & 84.20 & 89.71 \\

    \textbf{ICAL} & \textbf{69.06} & \textbf{85.16} & \textbf{90.61} \\
    
    \bottomrule
    	
    \end{tabular}
    
\end{table}
\subsection{Ablation Study} \label{sec:bt_eval}
Our research includes a series of ablation studies to corroborate the effectiveness of our proposed method. Presented in Table ~\ref{tb:ablation}, \textit{Initial Loss} refers to the cross-entropy loss computed from the discrepancy between the Transformer Decoder's direct output and the ground truth (the intact \LaTeX\ code). \textit{Fusion Loss} is the cross-entropy loss determined by comparing the combined outputs from both the Implicit Character Construction Module(ICCM) and Decoder—synergized through the Fusion Module—with the ground truth (the intact \LaTeX\ sequence). \textit{Implicit Loss} is calculated using a cross-entropy formula with adaptive weighting, which evaluates the discrepancies between the ICCM's output and the sequence of implicit characters. When \textit{Implicit Loss} is not applied, the ICCM module is also omitted, serving as a validation of ICCM's effectiveness.

Additionally, it should be noted that in the ablation experiments mentioned above, when utilizing \textit{Fusion Loss}, we employ the output of the Fusion Module for inference. Conversely, when \textit{Fusion Loss} is not implemented, inference is conducted directly using the output from the Transformer Decoder.

From the 4th and 5th rows of each dataset in the table, it is evident that compared to using only the \textit{Initial Loss}(1st row), which serves as the baseline, CoMER, \textit{Implicit Loss}(4th row) and \textit{the Fusion Loss} (5th row)  that we have introduced can both effectively enhance the model's recognition performance. We also designed experiments that exclusively utilize \textit{Fusion Loss} and \textit{Implicit Loss} (the 3rd row), from which it can be observed that, relative to the baseline approach, our method still manages to achieve a notable improvement in effectiveness.

Due to time limitations, the ablation study conducted on the HME100K dataset only used a single random seed for each experiment, and thus, the results are not averaged over multiple runs. This may introduce some variability in the reported performance, and we plan to address this limitation by conducting additional experiments with multiple seeds in future work for more robust evaluation.

\begin{table}[t]
\renewcommand\arraystretch{0.8}
    \centering
    \caption{Ablation study on the CROHME 2014/2016/2019 and HME100K test sets(in \%). It should be noted that, if \textit{Implicit Loss} is not applied,  the ICCM module is also not used, which serves to validate the effectiveness of the ICCM. Similarly, when \textit{Fusion Loss} is not implemented, inference is conducted directly using the output from the Transformer Decoder.}
    \label{tb:ablation}
    \vspace{0.5em}
    
    \begin{tabular}{c|c|c|c|c}
    
    \toprule
    
    \textbf{Dataset} & \textbf{Initial Loss} & \textbf{Fusion Loss} & \textbf{Implicit Loss} & \textbf{ExpRate} \\
    
    \midrule
    
    \multirow{6}{*} {CROHME 2014}
    & \cmark &  &  & 58.38 \\
    &  &\cmark  &  & 58.52 \\
    &  & \cmark & \cmark & 59.02 \\
    & \cmark &  & \cmark & 59.25 \\
    & \cmark & \cmark &  & 60.04 \\
    & \cmark & \cmark & \cmark & \textbf{60.63} \\
    \midrule
    \multirow{6}{*} {CROHME 2016}
    & \cmark &  &  & 56.98 \\
    &  &\cmark  &  & 57.04 \\
    &  & \cmark & \cmark & 58.13 \\
    & \cmark &  & \cmark & 57.80 \\
    & \cmark & \cmark &  & 58.44 \\
    & \cmark & \cmark & \cmark & \textbf{58.79} \\
    \midrule
    \multirow{6}{*} {CROHME 2019}
    & \cmark &  &  & 59.12 \\
    &  &\cmark  &  & 59.15 \\
    &  & \cmark & \cmark & 59.70 \\
    & \cmark &  & \cmark & 60.40 \\
    & \cmark & \cmark &  & 59.61 \\
    & \cmark & \cmark & \cmark & \textbf{60.51} \\
    \midrule
    \multirow{6}{*} {HME100K}
    & \cmark &  &  & 68.12 \\
    &  &\cmark  &  & 68.18 \\
    &  & \cmark & \cmark & 68.47 \\
    & \cmark &  & \cmark & 68.46 \\
    & \cmark & \cmark &  & 69.09 \\
    & \cmark & \cmark & \cmark & \textbf{69.25} \\
    
    \bottomrule
    	
    \end{tabular}
    \vspace{-1em}
\end{table}
\subsection{Inference Speed} \label{sec:infer_speed}
As shown in Table ~\ref{tb:infer_speed}, we have evaluated the inference speed of our method on a single NVIDIA 2080Ti GPU. Compared to the baseline model, our method has a modest increase in the number of parameters with a negligible impact on FPS, and there is also a slight increase in FLOPs.

\begin{table}[t]
    \centering
    \caption{Comparative Analysis of Parameters (Params), Floating-Point Operations (FLOPs), and Frames Per Second (FPS)}
    \label{tb:infer_speed}
    \setlength\tabcolsep{8pt}
    \renewcommand\arraystretch{1.2} 
    \begin{tabular}{c|c|c|c|c}
        \hline
        \textbf{Method} & \textbf{Input Image Size} & \textbf{Params (M)} & \textbf{FLOPs (G)} & \textbf{FPS} \\
        \hline
        CoMER & (1,1,120,800) & 6.39 & 18.81 & 2.484 \\
        ICAL & (1,1,120,800) & 7.37 & 19.81 & 2.394 \\
        \hline
    \end{tabular}
\end{table}

\subsection{Case Study} \label{sec:case_study}
\begin{figure}[htbp]
	\centering
	\includegraphics[scale=0.6]{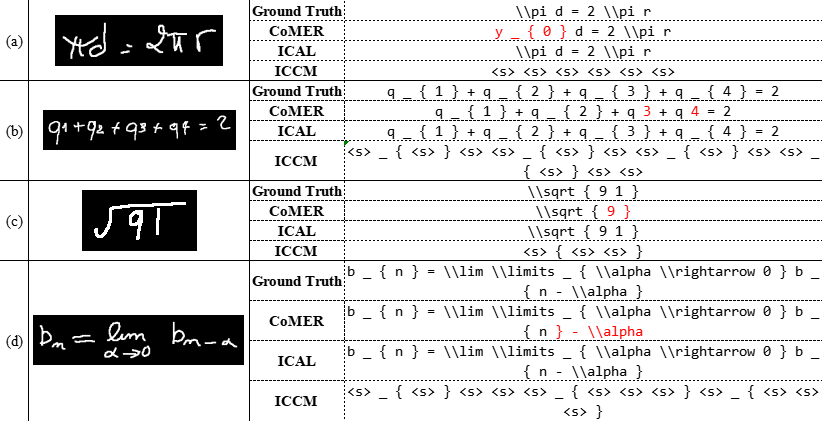}
	\caption{Case studies for the  Ground Truth and CoMER, ICAL methods. The red symbols represent incorrect predictions. 'ICCM' represents the implicit character sequence predicted by the ICCM module, where $\texttt{<s>}$ is the abbreviation for the $\texttt{<space>}$ token.}
	\label{fig:case_study}
\end{figure}

We provide several typical recognition examples to demonstrate the effectiveness of the proposed method, as shown in Fig.~\ref{fig:case_study}. Entries highlighted in red indicate cases where the model made incorrect predictions. 'ICCM' represents the implicit character sequence predicted by the ICCM module, where \verb|<s>| is the abbreviation for the \verb|<space>| token.

In Case (a), the baseline CoMER model incorrectly identified the first character, $\pi$, as \verb|y_{0}|. In contrast, the ICCM correctly determined that there were no implicit characters in the formula, as indicated on the fourth line of Group a. This accurate detection allowed for the correct \LaTeX\  sequence to be output by ICAL, as shown on the third line of Group a.

In Case (b), the ICCM’s prediction of the implicit character sequence (fourth line of Group b) was crucial. It enabled the ICAL method to correctly place both characters \verb|3| and \verb|4| in the subscript of \verb|q| (third line of Group b), unlike CoMER, which misidentified this relationship (second line of Group b).

Case (c) demonstrates that the ICCM's prediction of implicit characters can also alleviate the lack of coverage issue~\cite{tu2016modeling, zhang2017watch} . Here,  ICCM's prediction indicated that there should be two explicit characters, \verb|9| and \verb|1|, within \verb|{ }|,  and correspondingly, ICAL also successfully predicted these two characters. However, CoMER only predicted the character \verb|9|, and missed the character \verb|1|.

Moreover, Case (d) also effectively highlights how ICCM and its ability to predict implicit characters can enhance a model's understanding of the structural relationships within formulas.

\section{Conclusion}
In this paper, we propose a novel recognizer framework, ICAL, capable of leveraging global information in \LaTeX\ to correct the predictions of the decoder. Our main contributions are threefold: (1) We have designed an Implicit Character Construction Module(ICCM) to predict implicit characters in \LaTeX. (2) Additionally, we employ a Fusion Module to aggregate global information from implicit characters, thereby refining the predictions of the Transformer Decoder. We integrate these two modules into the CoMER model to develop our method, ICAL. (3) Experimental results demonstrate that the ICAL method surpasses previous state-of-the-art approaches, achieving expression recognition rate (ExpRate) of 60.63\%, 58.79\%, and 60.51\% on the CROHME 2014, 2016, and 2019 datasets, respectively, and an ExpRate of 69.06\% on the HME100K dataset.

\subsection*{Acknowledgements}
This work is supported by the projects of National Science and Technology Major Project (2021ZD0113301) and National Natural Science Foundation of China (No. 62376012), which is also a research achievement of Key Laboratory of Science, Technology and Standard in Press Industry (Key Laboratory of Intelligent Press Media Technology).
\newpage
%
%
%

\bibliographystyle{splncs04}
\bibliography{main}

\end{document}